\documentclass[10pt,letterpaper]{article}
\pdftrailerid{}
\usepackage[T1]{fontenc}
\usepackage[letterpaper,left=1.05in,right=1.05in,top=.70in,bottom=.77in,footskip=23pt]{geometry}
\usepackage{amsmath,amssymb,amsfonts}
\usepackage{newtxtext,newtxmath}
\usepackage[scaled=.94]{helvet}
\usepackage{microtype,graphicx,booktabs,array,multirow,float,placeins}
\usepackage[table]{xcolor}
\usepackage{tcolorbox}
\usepackage{caption}
\usepackage{pifont,url,xspace,soul,enumitem,tikz,tabularx,longtable,upquote}
\usepackage[normalem]{ulem}
\usepackage[numbers,sort&compress]{natbib}
\usepackage{titlesec,fancyhdr}
\usepackage[colorlinks,allcolors=linkblue,bookmarksnumbered,pdfdisplaydoctitle]{hyperref}
\definecolor{ink}{HTML}{151B2B}
\definecolor{linkblue}{HTML}{314D9A}
\definecolor{quiet}{HTML}{515969}
\definecolor{rulegray}{HTML}{CBD1DB}
\definecolor{coverink}{HTML}{1C2B34}
\definecolor{abstractfill}{HTML}{F1F4F6}
\definecolor{cvprblue}{rgb}{0.21,0.49,0.74}
\hypersetup{pdftitle={CaAP: Code as Adaptive Policies for Cold-Start Robot Manipulation},
pdfauthor={Pianran Huang, Ziyi Liu, Zhaoxin Luo, Qichen Tan, Jiajie Xu, He Huang, Min Zhang, Zhijun Li},
pdfsubject={Code-as-policy robot manipulation}}
\newif\ifdraftnotes \draftnotesfalse
\newcommand{\todo}[1]{}
\newcommand{\projnote}[1]{}

\titleformat{\section}{\Large\sffamily\bfseries\color{ink}}{\color{linkblue}\thesection}{.65em}{}
\titleformat{\subsection}{\large\sffamily\bfseries\color{ink}}{\thesubsection}{.65em}{}
\titleformat{\paragraph}[runin]{\normalsize\sffamily\bfseries\color{ink}}{\theparagraph}{.5em}{}
\titlespacing*{\section}{0pt}{2.1ex plus .6ex minus .3ex}{.65ex}
\titlespacing*{\subsection}{0pt}{1.65ex plus .4ex minus .25ex}{.45ex}
\newcommand{\aff}[1]{\textsuperscript{#1}}
\newcommand{\schoolaffiliation}[4]{%
  \mbox{\aff{#1}\,\href{#2}{%
    \raisebox{-.6pt}{#3}%
    \hspace{.28em}\textcolor{quiet}{#4}}}}
\DeclareCaptionFont{covercaption}{\fontsize{9.55}{11.7}\selectfont}
\IfFileExists{T1SourceSansThree-TLF.fd}%
  {\newcommand{\coversans}{\fontfamily{SourceSansThree-TLF}\selectfont}}%
  {\newcommand{\coversans}{\fontfamily{SourceSansPro-TLF}\selectfont}}
\fancypagestyle{cover}{\fancyhf{}\fancyfoot[C]{\small\color{quiet}\thepage}}
\begin{document}
\thispagestyle{cover}
\noindent\href{https://www.insightos.cn/}{\includegraphics[height=23.5pt]{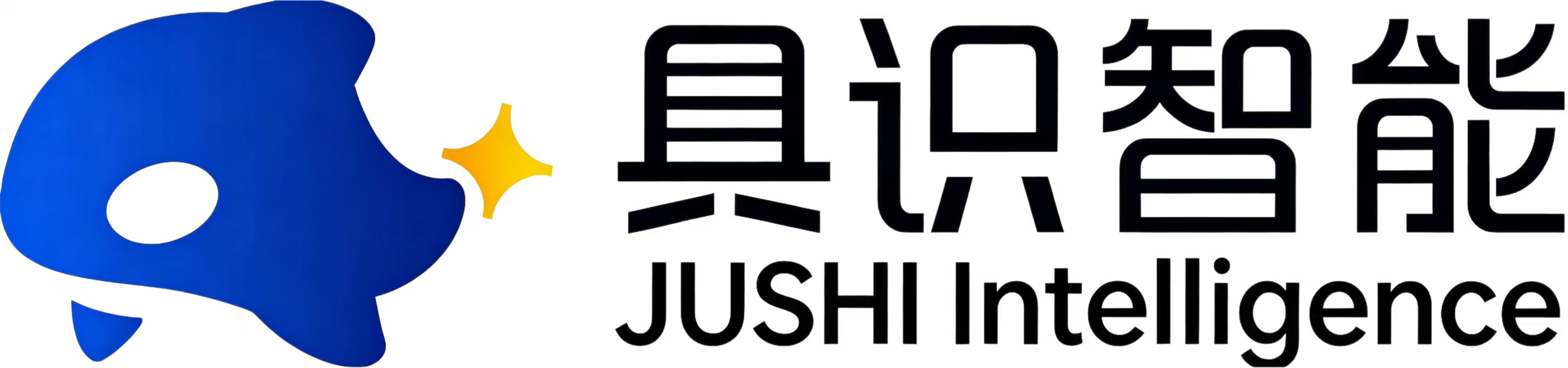}}\par
\vspace{16pt}
{\noindent\raggedright\hyphenpenalty=10000\coversans\fontseries{sb}\fontsize{18.05}{21.5}\selectfont
\color{coverink}CaAP: Code as Adaptive Policies for Cold-Start Robot Manipulation\par}
\vspace{7pt}{\color{rulegray}\hrule height .45pt}
\vspace{8pt}
{\noindent\coversans\fontseries{sb}\fontsize{9.6}{12}\selectfont\color{coverink}%
Pianran Huang\aff{1}, Ziyi Liu\aff{2,\textdagger}, Zhaoxin Luo\aff{1}, Qichen Tan\aff{2}, Jiajie Xu\aff{2}, He Huang\aff{2}, Min Zhang\aff{1}, Zhijun Li\aff{1,\textdagger}\par}
\vspace{3pt}
{\noindent\fontsize{9.25}{11.5}\selectfont\color{quiet}%
\schoolaffiliation{1}{https://en.hit.edu.cn/}{\includegraphics[height=7.2pt]{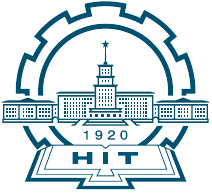}}{Harbin Institute of Technology}\qquad
\schoolaffiliation{2}{https://www.suda.edu.cn/}{\includegraphics[height=7.2pt]{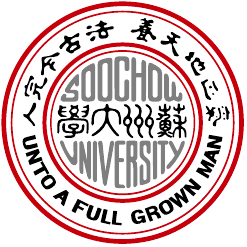}}{Soochow University}\par}
\vspace{3.5pt}
{\noindent\fontsize{9}{11}\selectfont\color{quiet}%
First author\enspace
\href{mailto:25S003016@stu.hit.edu.cn}{\nolinkurl{25S003016@stu.hit.edu.cn}} (Pianran Huang)\\
\aff{\textdagger}Corresponding authors\enspace
\href{mailto:liuziyi@suda.edu.cn}{\nolinkurl{liuziyi@suda.edu.cn}} (Ziyi Liu)\quad
\href{mailto:lizhijun_os@hit.edu.cn}{\nolinkurl{lizhijun_os@hit.edu.cn}} (Zhijun Li)\par}
\nointerlineskip
\vbox to 23pt{\vfil
  \hbox to \linewidth{%
    {\fontsize{10.5}{13}\selectfont
    \href{https://hollow2003.github.io/CaAP-ProjectPage/}{%
      \raisebox{-.65pt}{\includegraphics[height=9.25pt]{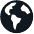}}%
      \hspace{3pt}Project Page}%
    \hspace{11pt}\textcolor{coverink}{\textbar}\hspace{11pt}%
    \href{https://github.com/hollow2003/CaAP}{%
      \raisebox{-.65pt}{\includegraphics[height=9.25pt]{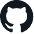}}%
      \hspace{3pt}Code}}%
    \hfil}%
  \vfil}
\begin{tcolorbox}[width=\linewidth,colframe=abstractfill,colback=abstractfill,
  boxrule=0pt,arc=6pt,outer arc=6pt,boxsep=0pt,left=13pt,right=13pt,
  top=10pt,bottom=11pt,before skip=0pt,after skip=0pt]
{\coversans\fontseries{sb}\fontsize{8.1}{10}\selectfont\color{quiet}\textls[130]{ABSTRACT}\par}
\vspace{4.5pt}
{\noindent\color{coverink}\fontsize{10.3}{12.8}\selectfont Code-as-policy methods enable robot manipulation without task-specific training, but they often achieve low zero-shot success and require extensive interaction to adapt to new tasks.
We present CaAP, a framework that improves both zero-shot performance and few-trial adaptation efficiency. 
Within a trial, Self-Adaptive Process Monitoring lets the agent specify when and how to observe execution, while Heritable Handoff deliberately suspends the program when its assumptions no longer hold, preserving the scene and program state for subsequent recovery.
Across trials, LoreCraft consolidates successes and failures into declarative lessons and reusable executable tools, enabling subsequent attempts to build on prior experience. 
On LIBERO-PRO, CaAP achieves 53.9\% average zero-shot success, about three times that of the strongest code-based baseline, CaP-Agent0 (18.0\%), and 10.3 percentage points above the strongest evaluated vision-language-action baseline. 
After 15 online adaptation trials per task, validation-based program selection raises the average success rate to 73.1\%.
This exceeds the 61.7\% reported by ASPIRE with Opus 4.6 while using about one-sixth of its rollouts and one-tenth of its cost. 
Experiments on a real bimanual mobile manipulator further demonstrate the applicability of CaAP to physical robot manipulation.
\par}
\end{tcolorbox}
\vspace{17pt}
{\captionsetup{font=covercaption,margin=6pt,labelfont={sf,bf,color=coverink}}\begin{figure}[H]
\centering
\includegraphics[width=.55\linewidth]{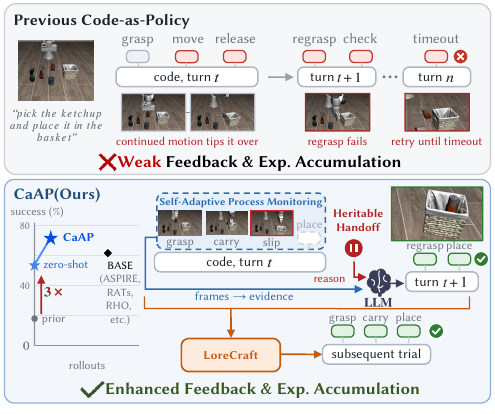}
\caption{Prior code-as-policy versus CaAP. \textbf{Top:} a prior agent runs each turn's code to the end, tips the object over and later turns cannot grasp the fallen object. \textbf{Bottom:} CaAP captures the uncertain motion, hands off before the next step, and the LLM writes turn $t{+}1$ from both; LoreCraft carries the lessons to later trials. \textbf{Bottom left:} success rate vs.\ rollouts (LIBERO-PRO, 60 tasks).}
\label{fig:teaser}
\end{figure}
}
\clearpage
\section{Introduction}

Robotic manipulation depends not only on planned actions, but also on what happens during their execution. An object can slip, a motion can be obstructed, or a visual check can misinterpret the scene. When a robot continues executing a program based on an incorrect assumption, subsequent actions can turn a recoverable problem into task failure. Recovering requires the robot to reconsider its actions without losing completed progress. For unfamiliar tasks, the robot must also learn from limited attempts rather than repeatedly making the same mistakes.

Recent language-model-based methods translate task instructions into executable robot programs using perception and control tools~\cite{liang2023codeaspolicies,singh2023progprompt}. They enable manipulation without task-specific training, but their performance remains sensitive to unfamiliar objects, instructions, and execution conditions. Methods such as Inner Monologue~\cite{huang2022inner} incorporate environmental feedback, while Code-as-Monitor~\cite{zhou2025codeasmonitor} and HyCodePolicy~\cite{liu2025hycodepolicy} introduce execution monitoring and diagnosis. Other approaches, including CaP-Agent0~\cite{liu2026capx} and ASPIRE~\cite{lu2026aspire}, improve generated programs through repeated interaction. Despite these advances, reliable execution and efficient adaptation remain difficult when observations are limited and each trial is costly.

We introduce \textit{\uline{C}ode \uline{a}s \uline{A}daptive \uline{P}olicies (CaAP)}, which give generated programs greater control over their execution. Rather than observing only at predefined points, \emph{Self-Adaptive Process Monitoring} lets the code choose which action intervals to record and how to sample them. A visual observer, termed \emph{Witness}, summarizes the collected evidence after the code block ends. When an assumption fails or requires further examination, \emph{Heritable Handoff} allows the code to deliberately return control with its reason and diagnostics. The next agent turn inherits the current scene, program state, and visual evidence to decide how to proceed. This allows the agent to reconsider its actions without discarding the progress already made.

Beyond a single execution, CaAP learns from previous trials through \emph{LoreCraft}. A failed trial can contain successful actions worth retaining, while a successful trial can include unnecessary steps. Learning from the overall outcome alone cannot distinguish them. LoreCraft uses execution records aligned with code blocks, API calls, and visual observations to identify useful experience. It maintains two complementary forms of knowledge. \emph{Lore} captures declarative lessons for planning and verification, while \emph{Craft} stores reusable executable tools. Both are updated between trials and made available to subsequent attempts. In this way, CaAP combines adaptive execution within a trial with experience learning across trials, improving zero-shot performance and adaptation with limited interaction.

Our main contributions are summarized as follows.

\begin{itemize}[leftmargin=0.6cm]
\item We introduce CaAP, a code-as-policy framework that improves zero-shot manipulation and few-trial adaptation through adaptive execution and experience reuse.

\item We develop Self-Adaptive Process Monitoring and Heritable Handoff, allowing generated programs to control observation and deliberately return control while preserving execution context.

\item We propose LoreCraft, which learns declarative lessons and reusable executable tools from execution-aligned evidence across successful and failed trials.

\item Experiments on LIBERO-PRO and a real bimanual mobile manipulator demonstrate improved zero-shot success and adaptation efficiency with fewer rollouts and lower computational cost.

\end{itemize}

\section{Related Work}

\paragraph{Robot policy generation.}
Robot manipulation policies can be learned directly from visual observations and instructions using vision-language-action (VLA) models~\cite{brohan2023rt2,kim2024openvla,black2024pi0,intelligence2025pi05,kim2025oft,zhao2025cotvla,jang2026ravla}, or generated by language models that compose perception and control tools into executable programs~\cite{aher2022saycan,liang2023codeaspolicies,gupta2023visprog,huang2023voxposer,kim2023demo2code,yao2023react}. Both approaches face challenges when task conditions differ from those previously encountered, as examined in LIBERO-PRO and Colosseum~\cite{zhou2026liberopro,morgan2026colosseum}. More recently, coding agents have enabled multi-turn program generation and revision. CaP-X~\cite{liu2026capx} benchmarks this setting and introduces CaP-Agent$0$, a training-free baseline. These advances extend robot control beyond fixed policies, making execution reliability and rapid adaptation increasingly important.

\paragraph{Execution feedback.}
Execution feedback helps robots assess actions and recover from unexpected outcomes. Inner Monologue~\cite{huang2022inner} and REFLECT~\cite{liu2023reflect} incorporate environmental feedback and execution histories into planning and failure analysis. Code-as-Monitor~\cite{zhou2025codeasmonitor} generates visual constraint monitors that trigger replanning, while Phoenix, RoboRepair, HyCodePolicy, and CodeTracer investigate execution monitoring, diagnosis, and program repair~\cite{xia2025phoenix,schlesinger2024roborepair,liu2025hycodepolicy,li2026codetracer}. These methods differ in their observation and recovery mechanisms. Our work focuses on giving generated code explicit control over which execution intervals to observe and when to yield control, while retaining the execution context for subsequent decisions.

\paragraph{Self-improvement.}
Language agents improve through reflection, experience extraction, and reusable workflows~\cite{shinn2023reflexion,zhao2024expel,wang2024awm,suzgun2025dynamiccheatsheet}, as well as executable skill libraries and program evolution~\cite{wang2023voyager,novikov2025alphaevolve}. In robotics, RoboCat~\cite{bousmalis2023robocat} learns from self-generated data, RATs~\cite{zhang2026rats} acquires skills through exploration, and RHO~\cite{elmaaroufi2026rho} evolves policy code. ASPIRE~\cite{lu2026aspire} iteratively debugs programs and accumulates reusable skills, while Act--Observe--Rewrite~\cite{kumar2026aor} revises controllers using execution feedback. These methods demonstrate different approaches to learning from interaction. CaAP focuses on few-trial adaptation, using evidence aligned with executed code to identify useful actions and retain both declarative lessons and reusable tools from successful and failed trials.

\begin{figure}[t]
    \centering
    \includegraphics[width=\textwidth]{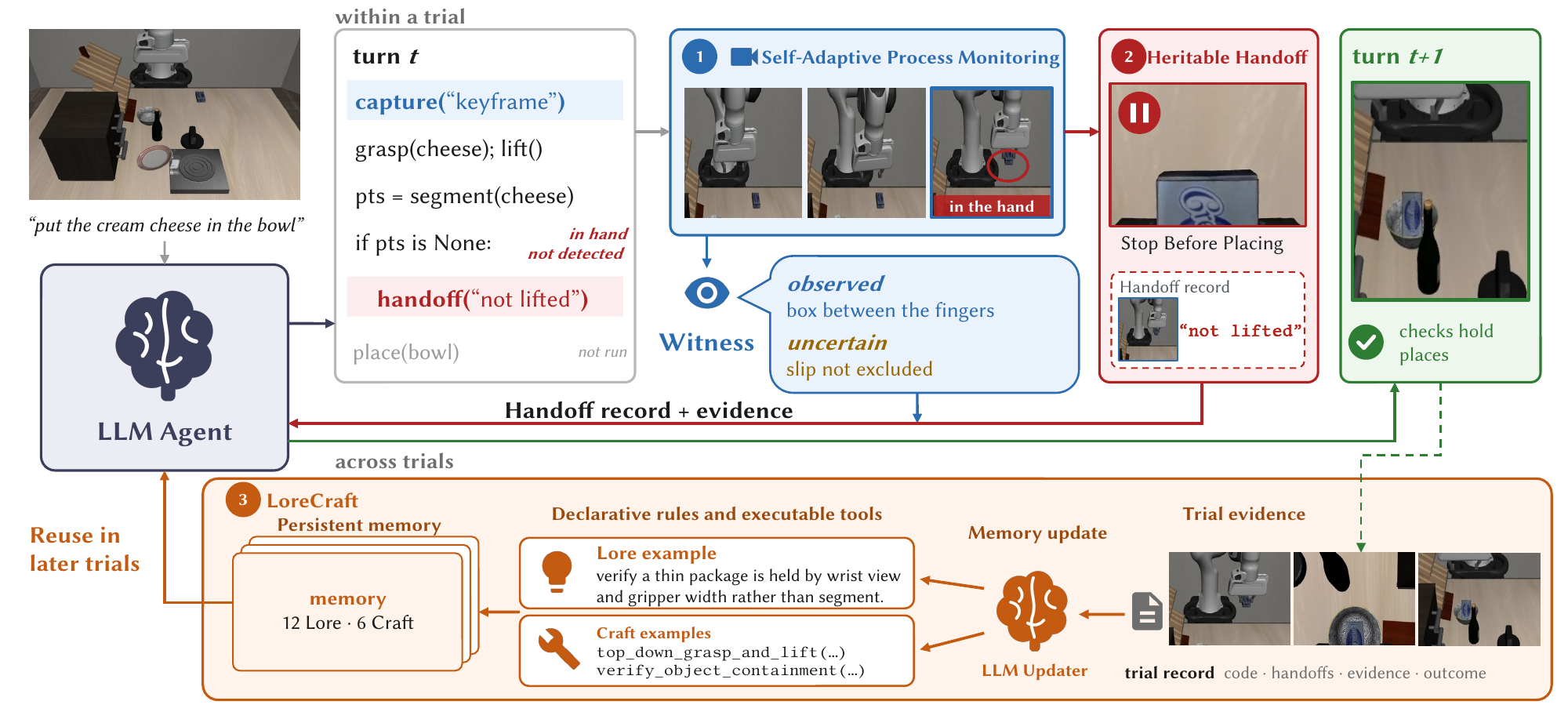}
    \caption{\textbf{CaAP overview.}
    Within a trial, generated code selects action intervals
    for visual monitoring and can deliberately hand off
    control when further decisions are needed.
    Witness evidence and handoff diagnostics are available
    to the next agent turn.
    Across trials, LoreCraft updates declarative lessons
    (Lore) and executable tools (Craft) from aligned
    execution records.
    The example shows a successful lift incorrectly judged
    as a failure by a segmentation check; handoff allows
    the next turn to verify the grasp differently and
    complete the placement.}
    \label{fig:method}
\end{figure}

\section{Preliminaries}
\label{sec:prelim}

\paragraph{Task and trials.}
We study language-instructed robotic manipulation in unseen scenes.
The agent uses fixed perception and control APIs without task-specific training, demonstrations, or prior rollouts.
Each task is attempted over multiple trials indexed by $i$, with the scene reset between trials.
Zero-shot performance is measured with empty memory on separate evaluation seeds, while the sequence of trials is used for adaptation from previous experience.

\paragraph{Within-trial execution.}
Each trial consists of agent turns indexed by $t$.
At each turn, the agent generates and executes a code block based on the current observation, persistent execution state, and accumulated evidence $L_{i,t}$.
The code performs robot actions, can specify which action intervals to observe, and can return control for reconsideration.
Execution state persists across turns, while collected visual evidence becomes available at the next turn.

\paragraph{Across-trial learning.}
Each trial produces an execution record $R_i$ containing executed code, API calls, and turn-aligned visual evidence.
The agent maintains memory $M_i$, consisting of declarative lessons and reusable executable tools, with $M_1=\varnothing$.
Memory remains fixed during each trial and is updated afterward by
\begin{equation}
    M_{i+1}=U(M_i,R_i,s_i),
    \label{eq:memory_update}
\end{equation}
where $U$ is the memory updater and $s_i\in\{0,1\}$ indicates trial success.
The outcome is available only to the updater after the trial ends.

\paragraph{Problem formulation.}
Given an unseen language-instructed manipulation task and a limited trial budget, we aim to develop reliable code-based policies without task-specific training.
The agent adapts its execution using accumulated evidence $L_{i,t}$ and available memory $M_i$, while the updater $U$ learns from trial records $R_i$ and outcomes $s_i$ to improve subsequent attempts.
Our objective is to improve task success on unseen initial states, both with initial memory $M_1=\varnothing$ for zero-shot execution and updated memory $M_i$, where $i>1$, after limited adaptation.

\section{Method}
\label{sec:method}

\paragraph{Overview.} CaAP separates decisions made during execution from
experience learned across trials (Fig.~\ref{fig:method}).
Within a trial, generated code selects when to collect
visual evidence and when to return control to the agent.
Between trials, LoreCraft updates reusable knowledge
from what was executed and observed.
All three mechanisms operate over the same perception
and control APIs.

\subsection{Self-Adaptive Process Monitoring}
\label{sec:monitor}

A generated program needs more than the final result
of an action to assess its execution.
CaAP lets the code mark an action interval for observation
using a pair of start and end calls.
The calls occur in the same code block and define a
\emph{capture window} $w=(\tau_s,\tau_e,m)$,
where $\tau_s$ and $\tau_e$ bound the interval and
$m$ specifies how its frames are sampled.
Closing a window records its endpoint without stopping
the remaining code.

\begin{figure}[t]
    \centering
    \includegraphics[width=.68\linewidth]{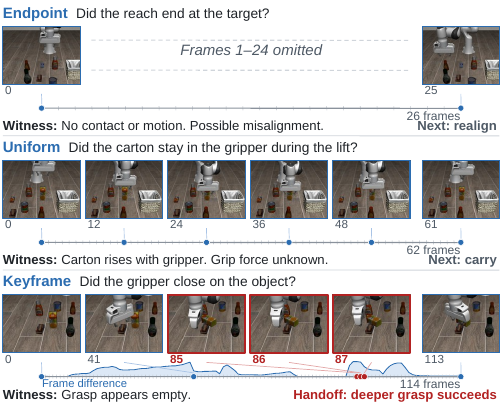}
    \caption{\textbf{Capture-window sampling.}
    Endpoint compares the beginning and end of an action,
    uniform follows its motion, and keyframe selects
    frames with large visual changes.
    In the keyframe example, the gripper closes on empty
    space; the next turn adjusts the grasp after handoff.}
    \label{fig:capture}
\end{figure}

The code chooses among three sampling modes.
\emph{Endpoint} retains the first and last frames
to compare states before and after an action.
\emph{Uniform} samples at equal intervals to follow
continuous motion.
\emph{Keyframe} retains the endpoints and selects
additional frames with the largest changes from their
predecessors, measured by mean absolute pixel differences
between downsampled frames.
This sampling is deterministic and requires no
visual-model calls.
Fig.~\ref{fig:capture} shows examples of the three modes.

The code determines where to observe and which mode
to use; the runtime controls the number of frames
under budget $B$.
We retain at most six frames per window, allow two
windows per block, and cap each trial at 64 frames.
If the remaining budget is insufficient, the runtime
returns fewer frames and records the shortfall.
Each window also records the turn, frame and
simulation-step indices, and the monitoring reason
supplied by the code.
The selected evidence can therefore be traced
to the corresponding action interval.

The visual observer, \emph{Witness} $V$, turns sampled
frames $S_m(w;B)$ into evidence
$e^{(w)}_{i,t}=V(S_m(w;B))$.
It describes visible motion, contact, object displacement,
collisions, and occlusions, and distinguishes observations
from uncertainty.
It does not infer task success from visual appearance alone.
Witness also compares the beginning and end of each
code block, including blocks without explicit capture
windows.

Evidence enters the ledger $L_{i,t+1}$ only after the
block exits, whether by normal return, handoff, or error.
The executing block cannot consume its own Witness report.
At the next turn, the agent receives the reports with
their code alignment and monitoring reasons.
Capture windows, sampled frames, and reports are retained
in an auditable manifest.

\subsection{Heritable Handoff}
\label{sec:handoff}

Observing an execution does not by itself prevent
the remaining program from acting on an incorrect
assumption.
CaAP therefore lets generated code call
\texttt{handoff(reason, diagnostics)} whenever
it needs to return control before running subsequent
statements.
The executor stops the current block and records
a \emph{handoff} exit rather than an execution error.
A block can thus finish normally (\emph{return}),
stop deliberately (\emph{handoff}), or terminate
with an uncaught exception (\emph{error}).
Unlike a traceback from an error, a handoff records
the condition that prompted the stop and the
information needed for reconsideration.

Handoff is useful after an observation that requires
interpretation, when an action precondition is not
satisfied, or before proceeding under an uncertain
result.
For example, a program unable to verify a grasp can
hand off before attempting placement.
No physical action is rolled back.
The next agent turn inherits the current scene,
the persistent variables, and the handoff reason and
diagnostics; it also receives the Witness evidence
from the completed block.
The agent then generates a new block based on this state.
Handoff returns control but does not itself diagnose
the failure or perform recovery.

Normal returns and runtime errors also preserve the
execution state, although the state after an error is
unverified.
The distinction is that handoff gives the program
an intentional, annotated boundary for requesting
another decision, rather than waiting for completion
or failure.

\subsection{LoreCraft}
\label{sec:lorecraft}

CaAP carries experience across trials in two forms.
\emph{Lore} stores declarative lessons for planning,
verification, failure recognition, and recovery.
\emph{Craft} stores reusable Python functions with
semantic parameters, observable pre- and postconditions,
failure behavior, and usage constraints.
Their documentation is provided to the agent, and
Craft implementations are loaded into the execution state.

After each trial, LoreCraft updates memory according
to Eq.~\eqref{eq:memory_update}.
Two updaters, $U_E$ for Lore and $U_T$ for Craft,
examine the execution record $R_i$ and outcome $s_i$.
The record includes code blocks, API calls that actually
ran (including window and handoff calls), and visual
keyframes aligned with each turn.
A vision model selects up to four keyframes from twelve
uniformly sampled candidates per turn.
These keyframes serve experience revision and are
distinct from the capture-window frames used for
process monitoring.
The Lore updater also receives earlier trial outcomes;
the Craft updater receives native API documentation
so it does not recreate an existing tool.

The central problem is deciding which parts of a trial
deserve to be retained.
A failed trial can contain a successful grasp or
transfer, and a successful trial can contain ineffective
actions.
LoreCraft therefore relates recorded API calls to
observed state changes, rather than treating the overall
outcome as a label for every step.
This permits a completed stage from a failed trial to
become a reusable Craft, while avoiding lessons
unsupported by execution evidence.
If the trial lacks footage aligned with its code,
neither updater revises memory.

\paragraph{Experience consolidation.}
Fig.~\ref{fig:method} illustrates how different parts
of an execution provide different learning signals.
The object is successfully lifted, but a segmentation
check incorrectly reports that it is missing.
The subsequent handoff allows the agent to verify
the grasp using other observations and complete the task.
The execution record separates the successful manipulation
from the unreliable verification step.
LoreCraft can use this evidence to retain a lesson about
grasp verification and extract reusable manipulation
procedures from the completed actions.
This avoids assigning the outcome of the entire trial
to every action it contains.
Each updater either proposes a complete memory revision
or returns \emph{no change}.
Proposed entries must be supported by the execution record.
Outcome labels are available only after the trial and
only to the updaters, never to the acting agent.
Lore cannot contain executable code or scene-specific
constants; scene-dependent quantities in Craft must be
parameters or derived from observations.
Revisions pass schema validation, with at most three
correction attempts before rejection.
Accepted versions are immutable, and each trial records
its memory versions and code hash for traceability.

\subsection{Perception and Control Toolkit}
\label{sec:toolkit}

CaAP uses a shared toolkit for text-prompted segmentation
with SAM 3~\cite{carion2025sam3}, visual pointing with
Molmo~\cite{deitke2025molmo}, RGB-D geometry through
OpenCV, numerical operations with NumPy, inverse
kinematics, and motion primitives.
The toolkit also exposes capture-window and handoff calls.
On the real robot, only the embodiment-specific interfaces
for the base, arms, and cameras are replaced;
the runtime and three CaAP modules remain unchanged.

\section{Experiments}
\label{sec:experiments}

We evaluate CaAP on zero-shot manipulation, adaptation from a few online trials, and interaction cost. Ablations isolate the contributions of within-trial control and cross-trial memory, and a real-robot experiment tests the same runtime outside simulation.

\subsection{Experimental Setup}
\label{sec:setup}

\paragraph{Models and benchmarks.}
The agent, Witness, and both LoreCraft updaters use DeepSeek-V4.1-Flash-1017. We evaluate on LIBERO-PRO~\cite{zhou2026liberopro}, built on LIBERO~\cite{liu2023libero} (robosuite/MuJoCo), at 20 Hz. Its Object, Goal, and Spatial suites each contain position (P) and task (T) perturbations, with 10 tasks per split and 60 tasks in total. All CaAP settings use the same limits on simulation steps and agent turns.

\paragraph{Evaluation protocol.}
Each task is evaluated zero-shot on seeds 1--50. Online adaptation uses 15 trials on seeds 51--65 in a fixed order. Programs obtained after trials 5, 10, and 15, together with the last successful trial, form the selection candidates. Each is evaluated on validation seeds 66--70. Candidates tied for the highest score are additionally checked on seeds 71--75. The best validation candidate is frozen and evaluated on seeds 1--50 without further model calls. Adaptation, selection, and evaluation thus use disjoint seeds.

\paragraph{Metrics and cost.}
We report \emph{ever-success} when the environment
success predicate holds at any simulation step,
and \emph{final success} when it holds at execution end.
A \emph{false completion} occurs when the agent
declares success without satisfying the task.
Ever-success is used as the trial outcome label. \textit{Adaptation} and \textit{interaction-cost} comparisons report ever-success unless noted otherwise. We count all target-task rollouts before evaluation, including online adaptation and validation, and report model costs where available.

\subsection{Zero-Shot Manipulation}
\label{sec:zeroshot}

\paragraph{LIBERO-PRO.}
Tab.~\ref{tab:zeroshot} compares CaAP with published VLA and code-based results. Without task-specific examples or prior rollouts, CaAP reaches 49.6\% final success and 53.9\% ever-success on average. Its final success exceeds CaP-Agent0~\cite{liu2026capx}, the strongest code-based baseline in the table (18.0\%), on all six splits. Under ever-success, CaAP exceeds the strongest VLA average in the table, Continuous Reasoning~\cite{wu2026continuous} (43.6\%), by 10.3 percentage points. It leads on five splits under that metric. Goal-T remains below PRTS~\cite{zhang2026prts}.

\begin{table}[t]
\centering
\fontsize{7.7}{9.2}\selectfont
\caption{Zero-shot success (\%) on six LIBERO-PRO splits. CaAP reports final and ever-success. VLA baselines are trained on LIBERO demonstrations, while code-based baselines use no target-task rollouts before evaluation. Code-based models are Claude Opus 4.6 (Opus 4.6), an ensemble of Gemini-3-Pro, GPT-5.2 and Claude Opus 4.5 (CaP-Agent0), and DeepSeek-V4.1-Flash-1017 (CaAP). Bold and underline mark the highest and second-highest values in each column.}
\label{tab:zeroshot}
\setlength{\tabcolsep}{3.4pt}
\begin{tabular*}{\columnwidth}{@{\extracolsep{\fill}}lccccccc@{}}
\toprule
\multirow{2}{*}{Method} & \multicolumn{2}{c}{Object} & \multicolumn{2}{c}{Goal} & \multicolumn{2}{c}{Spatial} & \multirow{2}{*}{Avg.} \\
\cmidrule(lr){2-3}\cmidrule(lr){4-5}\cmidrule(lr){6-7}
 & P & T & P & T & P & T & \\
\midrule
\multicolumn{8}{l}{\textit{VLA models}} \\
OpenVLA~\cite{kim2024openvla} & 0 & 0 & 0 & 0 & 0 & 0 & 0 \\
$\pi_0$~\cite{black2024pi0} & 0 & 0 & 0 & 0 & 0 & 0 & 0 \\
NORA~\cite{hung2025nora} & 0 & 0 & 0 & 0 & 0 & 0 & 0 \\
MolmoAct~\cite{lee2026molmoact} & 6.0 & 0 & 0 & 0 & 0 & 0 & 1.0 \\
X-VLA~\cite{zheng2026xvla} & 2.0 & 8.0 & 1.0 & 9.0 & 0 & 0 & 3.3 \\
Qwen3-VL-PI~\cite{starvla2025} & 0.2 & 0 & 1.2 & 8.8 & 14.4 & 0 & 4.1 \\
OpenVLA-OFT~\cite{kim2025oft} & 9.5 & 0 & 0 & 2.5 & 13.5 & 0 & 4.3 \\
VLA-Adapter~\cite{wang2026vlaadapter} & 0 & 8.0 & 0 & 12.0 & 0 & 49.0 & 11.5 \\
$\pi_{0.5}$~\cite{intelligence2025pi05} & 17.4 & 0.8 & 37.8 & 2.0 & 20.4 & 0.8 & 13.2 \\
ABot-M0~\cite{yang2026abotm0} & 9.0 & 9.8 & 5.6 & 10.6 & 13.4 & 54.8 & 17.2 \\
PRTS~\cite{zhang2026prts} & 36.0 & 8.8 & 19.0 & \textbf{33.8} & 26.8 & \underline{62.2} & 31.1 \\
Cont. Reasoning~\cite{wu2026continuous} & 54.5 & 27.0 & 36.0 & \underline{33.0} & 51.0 & 60.0 & 43.6 \\
\midrule
\multicolumn{8}{l}{\textit{Code-based methods}} \\
Opus 4.6~\cite{lu2026aspire} & 23.6 & 17.0 & 26.6 & 5.0 & 8.6 & 8.5 & 14.9 \\
CaP-Agent0~\cite{liu2026capx} & 21.8 & 18.2 & 25.6 & 16.8 & 11.8 & 14.0 & 18.0 \\
\textbf{CaAP (final)} & \underline{59.2} & \underline{53.2} & \underline{54.0} & 17.2 & \underline{51.6} & \underline{62.2} & \underline{49.6} \\
\textbf{CaAP (ever)} & \textbf{60.0} & \textbf{55.2} & \textbf{62.0} & 26.8 & \textbf{54.0} & \textbf{65.4} & \textbf{53.9} \\
\bottomrule
\end{tabular*}
\vspace{6pt}
\end{table}

\begin{figure}[t]
\centering
\includegraphics[width=\textwidth]{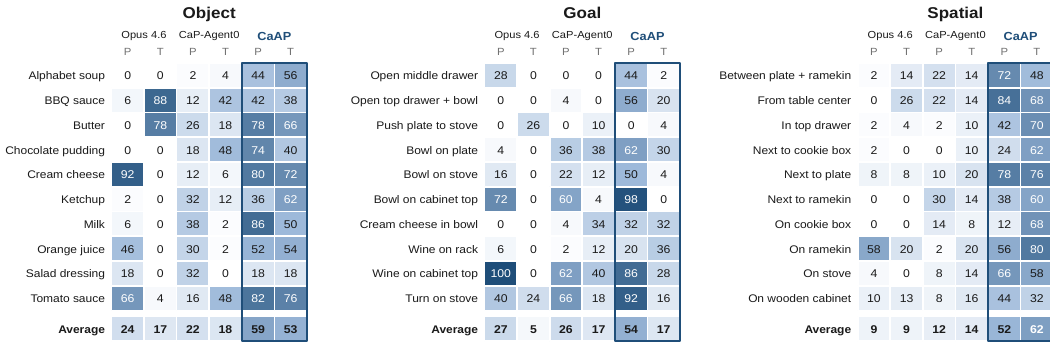}
\caption{Zero-shot success (\%) by LIBERO-PRO task. CaAP reports final success over 50 seeds per task. Baseline results are from ASPIRE~\cite{lu2026aspire} and CaP-X~\cite{liu2026capx}. P and T denote position and task perturbations.}
\label{fig:heatmap}
\end{figure}

Fig.~\ref{fig:heatmap} shows that the gains are not driven by a few tasks: CaAP exceeds both code-based baselines on most of the tasks. Goal-T is the hardest split, with 17.2\% final success compared with 62.2\% on Spatial-T, which motivates reporting adaptation results separately by perturbation type.

\paragraph{Recovery after goal disruption.}
Across all 3,000 zero-shot rollouts, CaAP reaches the goal at least
once in 1,617 executions and finishes successfully in 1,487. In 480
executions, a subsequent action temporarily disrupts an achieved goal
state; CaAP restores the goal in 350 of them (72.9\%). These recovered
executions account for 23.5\% of all final successes, showing that
execution feedback supports recovery even after intermediate progress
is partially undone. The remaining difference between ever-success and
final success is 4.3 percentage points on average, with larger gaps on
Goal-T and Goal-P. Because the simulator success predicate is hidden
from the acting agent, it must infer both recovery and completion from
visual observations and execution history. The results therefore
highlight both CaAP's ability to re-establish disrupted goal states and
the value of explicit goal-state checks before further actions.

\begin{figure}[t]
\centering
\includegraphics[width=\textwidth]{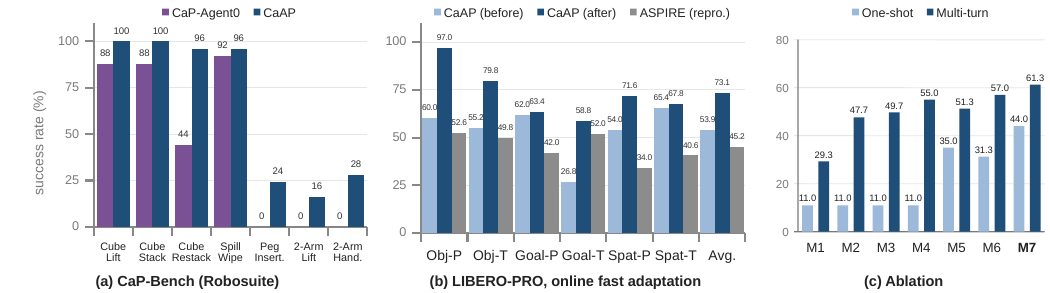}
\caption{Performance comparisons using DeepSeek-V4.1-Flash-1017. (a) CaP-Bench (Robosuite), seeds 1--25. (b) LIBERO-PRO before and after adaptation, including ASPIRE reproduced under a 34-rollout budget. (c) Component ablations on 30 tasks and seeds 1--10. M1 is Base. M2 adds Heritable Handoff to M1, M3 adds Self-Adaptive Process Monitoring to M1, and M4 adds both to M1. M5 and M6 add Lore and Craft, respectively, to M4. M7 adds both to M4 to give full CaAP. The memory variants use the selected Lore and Craft learned on separate adaptation seeds.}
\label{fig:bars}
\end{figure}

\paragraph{CaP-Bench.}\label{sec:capbench}
We further test seven Robosuite tasks from CaP-Bench~\cite{liu2026capx}, with no task memory and at most 10 agent turns. For a model-matched comparison, we rerun CaP-Agent0 using its released implementation with DeepSeek-V4.1-Flash-1017 and its nine-query ensemble procedure; CaAP uses one query per turn. Both are evaluated on seeds 1--25. CaAP averages 65.7\% success versus 44.6\% for CaP-Agent0 (Fig.~\ref{fig:bars}(a)). It scores 24\%, 16\%, and 28\% on Peg Insertion, Two-Arm Lift, and Two-Arm Handover, where the reproduced CaP-Agent0 records no success; on Cube Restack, it reaches 96\% versus 44\%.

\subsection{Few-Trial Online Adaptation}
\label{sec:online}

LoreCraft updates memory after each of the 15 online trials. Starting from empty task memory, it uses execution records from successful and failed attempts to consolidate declarative lessons into Lore and reusable procedures into Craft. Subsequent trials use this accumulated experience to guide planning, verification, and recovery.

\paragraph{Few-trial adaptation gains.}
Validation-based selection turns the experience from 15 online trials
into a frozen program that can be deployed without further model calls.
The selected programs achieve 73.1\% ever-success on held-out evaluation
seeds, improving the zero-shot average of 53.9\% by 19.2 points
(Fig.~\ref{fig:bars}(b)). Performance improves across all six splits:
Object-P rises from 60.0\% to 97.0\%, while Goal-T, the most challenging
zero-shot split, rises from 26.8\% to 58.8\%. Selection uses about 19
validation rollouts per task and no model calls, giving an average total
of 34 target-task rollouts including adaptation. Thus, a small amount of
online experience produces a substantially stronger program whose
evaluation-time execution is fully model-free.

\paragraph{Learned experience.}
After 15 trials, memory contains on average 10.2 Lore entries and 5.5 Craft tools, with only 4.8 and 3.7 revisions per task, respectively. Verification (30\%) and recovery (18\%) are the largest Lore categories, followed by procedure (10\%). Craft entries mainly cover perception, manipulation, geometry, and verification. Among the 370 inspected online programs from 56 tasks with Craft available, every program invokes at least one Craft tool. Thus, memory changes selectively rather than after every trial, and the learned tools are used during subsequent execution.

\subsection{Interaction Efficiency}
\label{sec:interaction}

\begin{figure}[!t]
\centering
\includegraphics[width=.62\linewidth]{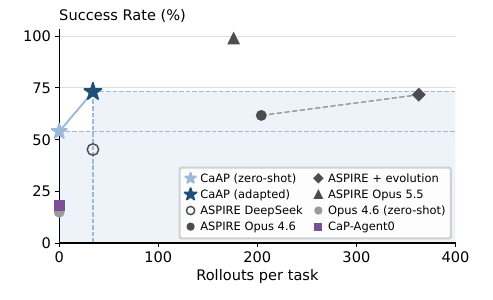}
\caption{LIBERO-PRO ever-success versus target-task rollouts before evaluation. Except for our budget-matched ASPIRE reproduction, baseline results are taken from the respective papers. Dashed lines indicate CaAP before and after adaptation.}
\label{fig:tradeoff}
\end{figure}

We compare success rate against the number of target-task rollouts before evaluation (Fig.~\ref{fig:tradeoff} and Tab.~\ref{tab:interaction}). RATs~\cite{zhang2026rats} acquires skills through a play phase, RHO~\cite{elmaaroufi2026rho} searches policy programs on training tasks, and ASPIRE~\cite{lu2026aspire} iteratively debugs generated programs. Their training or interaction protocols differ, so we report both the published results and a budget-matched ASPIRE reproduction.

With 34 rollouts and approximately \$2.3 per task, CaAP achieves 73.1\% ever-success. Published ASPIRE with Opus 4.6 achieves 61.7\% after 203.6 debugging rollouts and approximately \$22.9 per task. Its debugging-plus-evolution variant reaches 71.7\% with additional search. These are cross-model comparisons; ASPIRE with the stronger Opus 5.5 achieves 98.9\% using 175.9 rollouts per task. Under the same DeepSeek model and an approximately 34-rollout cap, our reproduction of ASPIRE reaches 45.2\%, versus 73.1\% for CaAP (Fig.~\ref{fig:bars}(b)). The average difference is 27.9 points, with CaAP ahead on all six splits. The main efficiency advantage over the published Opus 4.6 configuration is fewer target-task rollouts and lower reported cost.

\subsection{Ablation Study}
\label{sec:ablation}

Fig.~\ref{fig:bars}(c) evaluates the same five tasks (0, 1, 3, 5, and 7) from each of the six LIBERO-PRO splits, on seeds 1--10. The configurations share the DeepSeek model, perception and control APIs, Witness, and action-turn budget. In \emph{one-shot} generation, one program is generated on seed 1 and then frozen for seeds 1--10. In \emph{multi-turn} generation, the agent can revise code across turns within each trial. From Base (M1), we add Heritable Handoff (M2), Self-Adaptive Process Monitoring (M3), or both (M4). M5--M7 add Lore, Craft, or both to M4, using memory from the trial selected on the validation set. The memory was learned on seeds 51--65, disjoint from the ablation seeds.

\paragraph{Monitoring and handoff.}
In multi-turn generation, Base reaches 29.3\%. Adding Heritable Handoff raises success to 47.7\%, and adding Self-Adaptive Process Monitoring raises it to 49.7\%. Enabling both reaches 55.0\%, 25.7 points above Base. In one-shot generation, M1--M4 share the same frozen program and each scores 11.0\%.

The joint configuration exceeds the stronger single-module setting
by 5.3 points under the same model and action-turn budget.
Self-Adaptive Process Monitoring supplies evidence about an action
interval, and Heritable Handoff lets the agent use that evidence
to revise the remaining code.

\begin{table}[t]
\centering
\small
\caption{LIBERO-PRO success rate (\%) and target-task interaction before evaluation. Published methods use different backbone models and adaptation budgets; the bottom two rows compare ASPIRE and CaAP under the same DeepSeek model and approximately matched rollout budgets.}
\label{tab:interaction}
\resizebox{\textwidth}{!}{%
\begin{tabular}{lllccccccccc}
\toprule
\multirow{2}{*}{Method} & \multirow{2}{*}{Model} & \multirow{2}{*}{Interaction before evaluation} & \multicolumn{2}{c}{Object} & \multicolumn{2}{c}{Goal} & \multicolumn{2}{c}{Spatial} & \multirow{2}{*}{Avg.} & \multirow{2}{*}{Rollouts / task} & \multirow{2}{*}{Cost / task} \\
\cmidrule(lr){4-5}\cmidrule(lr){6-7}\cmidrule(lr){8-9}
 & & & P & T & P & T & P & T & & & \\
\midrule
RATs$^{1}$ & Gemini-3.1-Pro & play in the target environment & 61.0 & 63.0 & 43.0 & 36.0 & 29.0 & 31.0 & 43.8 & -- & -- \\
RHO & not stated & 63-generation search on unperturbed LIBERO & 12.2 & 37.2 & 50.6 & 55.6 & 53.0 & 61.6 & 45.0 & -- & -- \\
ASPIRE & Claude Opus 4.6 & debugging & 97.8 & 95.4 & 46.6 & 41.8 & 41.6 & 46.8 & 61.7 & 203.6 & \$22.9 \\
ASPIRE & Claude Opus 4.6 & debugging + evolution & 97.8 & 95.4 & 80.8 & 45.4 & 51.0 & 59.6 & 71.7 & 203.6 + search & \$22.9 + \$89.1$^{*}$ \\
ASPIRE & Claude Opus 5.5 & debugging & 98.2 & 100.0 & 97.0 & 99.4 & 99.4 & 99.2 & 98.9 & 175.9 & \$10.8 \\
\midrule
ASPIRE$^{2}$ & DeepSeek-V4.1-Flash-1017 & debugging, 34 rollouts & 52.6 & 49.8 & 42.0 & 52.0 & 34.0 & 40.6 & 45.2 & 34 & -- \\
\textbf{CaAP} & DeepSeek-V4.1-Flash-1017 & 15 online trials + validation & \textbf{97.0} & \textbf{79.8} & \textbf{63.4} & \textbf{58.8} & \textbf{71.6} & \textbf{67.8} & \textbf{73.1} & \textbf{34} & \textbf{\$2.3} \\
\bottomrule
\end{tabular}}
\par\vspace{3pt}
{\footnotesize\raggedright -- : not reported. Opus 4.6 and 5.5 results are from ASPIRE~\cite{lu2026aspire} (arXiv v2, Tables 10, 11, 20 and 21). $^{1}$ Evaluated on 10 initial states per task instead of 50. $^{2}$ Our reproduction, with the rollout cap enforced by the harness. $^{*}$ The search runs on the 31 tasks below 90\% success; its cost is per searched task.\par}
\end{table}

\begin{figure}[t]
\centering
\includegraphics[width=\textwidth]{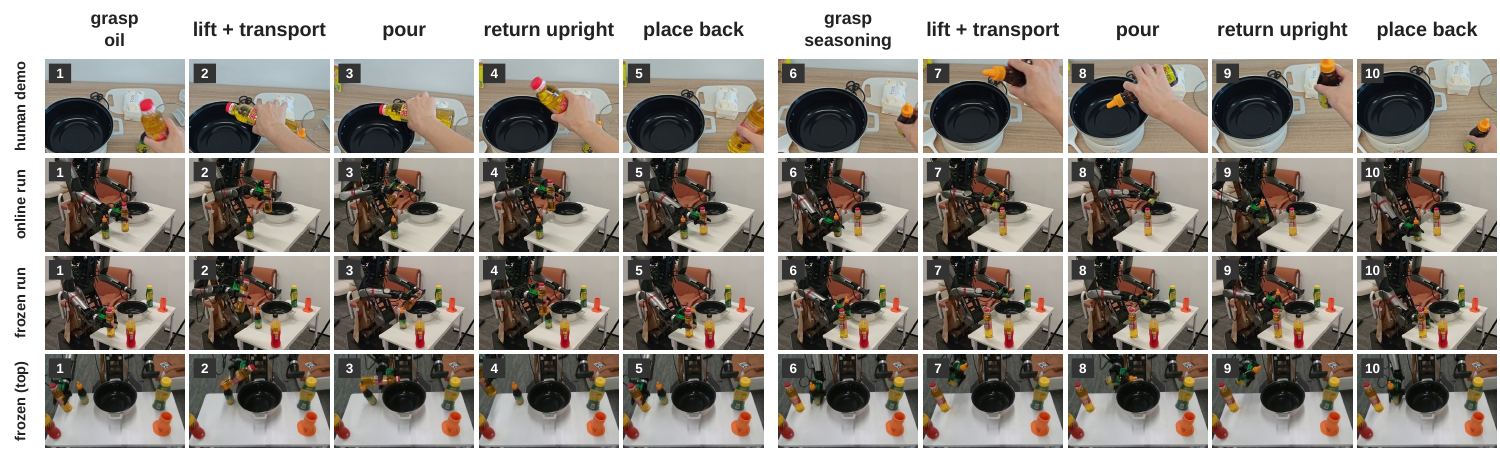}
\caption{Real-robot multi-stage pouring on a bimanual mobile manipulator. The task instruction is obtained from a human demonstration video (top) and specifies that the robot pour oil before adding seasoning. The online execution forms a program (second row). After freezing, the same program completes the ordered task without further model calls despite rearranged object positions and additional distractor objects (bottom rows).}
\label{fig:realrobot}
\end{figure}

\paragraph{Lore and Craft.}
Both forms of memory help, and they are complementary. In one-shot generation, where the program is frozen and every gain must come from code generated with memory, Lore alone raises success from 11.0\% to 35.0\% and Craft alone to 31.3\%; together they reach 44.0\%, four times M4. In multi-turn generation, where the agent can already repair many errors from execution feedback, M4 reaches 55.0\% and the combined memory raises it to 61.3\%, the best result in both modes. Lore contributes most when paired with Craft: on top of Craft, it adds 12.7 points in one-shot and 4.3 points in multi-turn generation, whereas lessons without the corresponding verified tools (51.3\%) do not improve over M4 in multi-turn generation.

\subsection{Real-Robot Evaluation}
\label{sec:vis}

We deploy the same runtime on a bimanual mobile manipulator, replacing
only the embodiment-specific interfaces. The task instruction is
obtained from a human demonstration video and requires the robot to pour
oil first and then add seasoning, returning each container after use.
Fig.~\ref{fig:realrobot} shows the online run and the resulting program
after it is frozen. Without further model calls, the frozen program
follows the demonstrated order and completes the task when the
containers are moved to new positions and additional distractor objects
are introduced. This transfer shows that the generated program does not
depend on a fixed scene layout and can re-observe the current scene to
ground each stage of the ordered task.

Each object sequence in Fig.~\ref{fig:realrobot} includes grasping,
transport, pouring, returning upright, and placement.
During the frozen run, the program observes the seasoning container
again after the oil sequence, updating its next grasp target from
the current scene.

\section{Conclusion}
\label{sec:conclusion}
We presented CaAP (Code as Adaptive Policies), a framework for cold-start deployment and fast adaptation to new tasks that treats execution itself as a learning signal. Within a trial, Self-Adaptive Process Monitoring exposes anomalies during execution, and Heritable Handoff stops in time and returns control to the agent. Across trials, LoreCraft consolidates the outcome of each trial into reusable lessons and tools. On LIBERO-PRO, without any prior interaction with the target environment, CaAP achieves 53.9\% average zero-shot success, above prior zero-shot code-based methods and vision-language-action models. After 15 online trials and selection, it reaches a 73.1\% success rate with 34 rollouts per task, about one-sixth of those used by ASPIRE with Opus 4.6 (61.7\%), and 27.9 points above ASPIRE under the same model and rollout budget. The same runtime also completes a multi-stage pouring task on a real bimanual mobile manipulator, where the frozen program transfers to a rearranged scene without further model calls. We hope this work moves code-as-policy methods from generating every program from scratch toward learning from execution.

\par\FloatBarrier\medskip\phantomsection\clearpage\phantomsection\addcontentsline{toc}{section}{Supplementary Material}
{\sffamily\footnotesize\color{quiet}CaAP\par}\vspace{3pt}
{\LARGE\bfseries\color{ink}Supplementary Material\par}\vspace{8pt}
\setcounter{table}{0}\setcounter{figure}{0}
\renewcommand{\thetable}{S\arabic{table}}\renewcommand{\thefigure}{S\arabic{figure}}

\appendix
\newsavebox{\suppcodebox}

\section{Implementation Details}
\label{sec:supp-impl}

\paragraph{Runtime.}
Each trial runs in a persistent Python namespace. At every agent turn the language model either returns one code block, which the executor runs in that namespace, or declares the task finished. Variables, helper functions and cached measurements defined in earlier blocks remain available to later blocks. The prompt for each turn contains all previously executed code blocks, the console history and the Witness reports. The agent never receives privileged simulator state, the success predicate or the task-completion signal; these are hidden by the environment wrapper and are used only for evaluation. Of this information, only the success label is passed to the LoreCraft updaters, after the trial ends.

\paragraph{Budgets.}
Tab.~\ref{tab:supp-budget} lists the limits shared by all CaAP settings on LIBERO-PRO. An action round is a code block that issues at least one control call (\texttt{goto\_pose}, \texttt{move\_to\_joints}, \texttt{goto\_home\_\allowbreak joint\_position}, \texttt{open\_gripper} or \texttt{close\_gripper}); several control calls in the same block count as one round. Perception, geometry, inverse kinematics, capture-window and handoff calls do not consume action rounds, so the agent can spend turns on observation when needed. There is no wall-clock limit on a trial; trials are bounded by the action-round and simulation-step budgets.

\begin{table}[!htbp]
\centering
\footnotesize
\caption{Runtime settings used for all CaAP results on LIBERO-PRO. The agent, Witness and both updaters use the same language model.}
\label{tab:supp-budget}
\setlength{\tabcolsep}{4pt}
\begin{tabular*}{\linewidth}{@{\extracolsep{\fill}}ll@{}}
\toprule
Setting & Value \\
\midrule
\multicolumn{2}{@{}l}{\textit{Model}} \\
Language model & DeepSeek-V4.1-Flash-1017 \\
Queries per agent decision & 1 (no ensemble) \\
Retries for a malformed response & 2 \\
\midrule
\multicolumn{2}{@{}l}{\textit{Execution}} \\
Control frequency & 20\,Hz \\
Simulation steps per trial & $\le$ 8{,}000 (400\,s) \\
Action rounds per trial & $\le$ 20 \\
Cameras & agent view and wrist, RGB-D \\
\midrule
\multicolumn{2}{@{}l}{\textit{Capture windows}} \\
Frames per window & $\le$ 6 \\
Windows per code block & $\le$ 2 \\
Frames per trial & $\le$ 64 \\
Sampling modes & endpoint, uniform, keyframe \\
\bottomrule
\end{tabular*}
\end{table}

\paragraph{Perception and control toolkit.}
All CaAP settings expose the same API (Tab.~\ref{tab:supp-api}). The same toolkit is used in zero-shot evaluation, online adaptation and all ablations.

\begin{table}[t]
\centering
\small
\caption{Functions available to generated code on LIBERO-PRO.}
\label{tab:supp-api}
\begin{tabularx}{\textwidth}{@{}l>{\raggedright\arraybackslash}X@{}}
\toprule
Group & Functions \\
\midrule
Observation & \texttt{get\_observation} (RGB, depth, intrinsics and camera pose for each camera; robot state) \\
Perception & \texttt{segment\_sam3\_text\_prompt}, \texttt{segment\_sam3\_point\_prompt} (SAM~3~\cite{carion2025sam3}); \texttt{point\_prompt\_molmo} (Molmo~\cite{deitke2025molmo}); \texttt{clean\_mask\_component}, \texttt{require\_valid\_perception} \\
Geometry & \texttt{depth\_to\_point\_cloud}, \texttt{mask\_to\_world\_points}, \texttt{pixel\_to\_world\_point}, \texttt{transform\_points}, \texttt{get\_oriented\_bounding\_box\_from\_3d\_points}, \texttt{subsample\_point\_cloud}, \texttt{filter\_noise}, \texttt{rotation\_matrix\_to\_quaternion}, \texttt{decompose\_transform}, \texttt{interpolate\_segment}, \texttt{normalize\_vector}, \texttt{mask\_geometry\_3d}, \texttt{select\_instance\_between}, \texttt{compute\_compensated\_place\_xy} \\
Kinematics and control & \texttt{solve\_ik}, \texttt{goto\_pose}, \texttt{move\_to\_joints}, \texttt{goto\_home\_joint\_position}, \texttt{open\_gripper}, \texttt{close\_gripper} \\
Process monitoring & \texttt{start\_visual\_recording}, \texttt{end\_visual\_recording} (Sec.~\ref{sec:supp-monitor}) \\
Handoff & \texttt{yield\_to\_next\_turn} (Sec.~\ref{sec:supp-monitor}) \\
\bottomrule
\end{tabularx}
\end{table}

\section{Process Monitoring, Witness and Handoff Interfaces}
\label{sec:supp-monitor}

\paragraph{Capture windows.}
A capture window is opened and closed by
\begin{center}\small
\texttt{start\_visual\_recording(mode='uniform', reason='')}\\ and \texttt{end\_visual\_recording(reason='')}.
\end{center}
Both calls must occur in the same code block. A nested start, an end without a start, or a window left open at the end of a block is reported as a protocol error. Only frames recorded between the two calls are eligible. Closing a window records its end point and does not stop the block. The code chooses the mode; the runtime chooses how many frames to keep within the budget of Tab.~\ref{tab:supp-budget} and returns fewer frames when the interval or the remaining budget is smaller. The calls do not move the robot and do not consume an action round.

\paragraph{Sampling modes.}
For an interval of $N$ frames and a budget of $k$ frames:
\begin{itemize}[leftmargin=*,itemsep=1pt,topsep=2pt]
  \item \emph{Endpoint} keeps frames $1$ and $N$.
  \item \emph{Uniform} keeps $k$ frames at evenly spaced indices from $1$ to $N$.
  \item \emph{Keyframe} keeps frames $1$ and $N$, then adds the frames with the largest change from their predecessors until $k$ frames are selected. The change of frame $j$ is the mean absolute difference between frames $j$ and $j{-}1$ after spatial downsampling by a factor of 8. The selected frames are returned in temporal order.
\end{itemize}
Selection is deterministic and uses no model calls.

\paragraph{Witness.}
Witness is the same language model called with images. It is used in three places: (i) a description of the initial scene before the first turn, (ii) a comparison of the first and last frames of every code block, in both cameras, and (iii) a description of the frames from each closed capture window together with the sampling mode and the reason supplied by the code. For capture windows, Witness is asked to describe only visible motion, contact, object displacement, placement error, collision and occlusion, to separate observation from uncertainty, and not to treat visual appearance as authoritative task-completion truth; its report is advisory evidence, not a success oracle. For the block comparison, Witness describes the change between the states before and after the block with the goal in mind. All reports are appended to the next agent prompt with the turn index, the code block and the monitoring reason; the agent alone decides whether to continue or declare completion.

\paragraph{Handoff.}
Heritable Handoff is exposed to generated code as
\begin{center}\small
\texttt{yield\_to\_next\_turn(reason="", **diagnostics)},
\end{center}
written \texttt{handoff(reason, diagnostics)} in the main paper. The call stops the current block immediately and records a handoff exit. It does not mean that the task is finished. The reason and the JSON-serializable diagnostics are passed to the next turn together with the unchanged scene and namespace. A block can therefore end in one of three ways: normal return, handoff, or an uncaught exception. Normal returns and exceptions also preserve the namespace.

\begin{table}[t]
\centering
\small
\caption{A Lore item (Object-P task~2) and a Craft tool (Object-P task~0), abridged.}
\label{tab:supp-memory-example}
\begin{tabularx}{\textwidth}{@{}lX@{}}
\toprule
\multicolumn{2}{@{}l}{\textbf{Lore} (verification): after lifting, distinguish an empty segmentation from an empty grasp} \\
\midrule
When & Re-segmentation of the target after lifting fails or is unreliable. \\
Recommendation & Do not conclude an empty grasp at once. Re-segment over the next 2--3 frames without color constraints, check whether the target mask appears near the gripper and moves with it, and read the gripper aperture. If the target moves with the gripper and the aperture is not fully closed, treat the grasp as held and continue. If the gripper occludes the target in the main view, use the wrist camera or the change in gripper width instead of repeating the same segmentation. \\
Verification & Across consecutive frames, the target stays fixed relative to the gripper and above the original table height. \\
Recovery & If the target remains on the table and the aperture is near fully closed, treat it as an empty grasp: open the gripper, return home, relocalize and retry. \\
Evidence & Six online trials (four successes, two failures) with cited keyframes; e.g., in one success re-segmentation returned nothing after lifting, while the frames show the object carried by the gripper. \\
Limitations & Requires several frames; the lift threshold depends on object size and must not be a fixed constant. \\
\midrule
\multicolumn{2}{@{}p{\textwidth}@{}}{\raggedright\textbf{Craft} (manipulation): \texttt{monitored\_lift(goto\_pose\_fn,} \texttt{measure\_gripper\_fn,} \texttt{quaternion,} \texttt{xy,} \texttt{z\_ladder,} \texttt{min\_grip=0.30,} \texttt{destination\_xy=None, \ldots)}} \\
\midrule
Summary & Lifts a closed gripper through a ladder of heights and reads the gripper aperture after each step; optionally transports horizontally in interpolated segments with the same check. Stops as soon as the aperture falls below the threshold and returns the failed phase with the full trace. \\
Preconditions & The gripper is closed on an object; the height ladder is increasing; a destination requires an interpolation function and a positive step. \\
Returns & \texttt{lifted}, \texttt{transported}, \texttt{grip\_lost}, \texttt{motion\_error} or \texttt{measurement\_error}, with the commanded positions and aperture readings of every step. \\
Non-goals & Does not align the grasp, place or release, verify placement, plan around obstacles or report task success. \\
\bottomrule
\end{tabularx}
\end{table}

\section{LoreCraft Details}
\label{sec:supp-lorecraft}

\paragraph{Trial record.}
After each trial, the record passed to the updaters contains the executed code blocks, the API calls that actually ran (including capture-window and handoff calls), the outcome label, and per-turn visual keyframes. For each turn, twelve frames are sampled uniformly from the turn video, and a vision call selects at most four of them as keyframes; images are resized to at most 512 pixels on the longer side. These keyframes are used only for memory revision and are separate from the capture-window frames used during execution. If a trial has no footage aligned with its code, neither updater revises memory.

\paragraph{Lore.}
Lore is a library of declarative items. Each item records when it applies, a recommendation, how to verify it, a knowledge type (perception, affordance, procedure, alignment, motion, failure, recovery or verification), a valence (positive, negative or mixed), a confidence level, retrieval tags, and the rollout and frame indices that support it. Lore must not contain executable code, task-completion oracles or scene-specific coordinates, orientations or constants.

\paragraph{Craft.}
Craft is a library of Python functions. Each entry stores the signature and implementation, a summary, preconditions, postconditions, failure behavior, non-goals, usage examples and anti-examples, and composition rules. Scene-dependent quantities must be function parameters or derived from observations. A capability inventory lists the fixed primitives of Tab.~\ref{tab:supp-api}; the Craft updater may call them but must not copy, rename or re-implement them.

\paragraph{Update rules.}
Both updaters run after every trial, at temperature~0. Each either returns \texttt{no\_change} or a complete replacement library, in which items are added, modified or deleted. Unaffected items are preserved with stable identifiers, and an item is deleted only when the record contains contradictory, harmful or redundancy evidence; absence from one trial is not evidence for deletion. The outcome label describes only the full task: a failed trial may contain successful phases and a successful trial may contain bad attempts, so local phases are judged from the code, the API records and the keyframes. A Craft tool is credited only when the record shows that it was actually called. A replacement is checked against the schema; on failure the updater is asked to correct it, with at most three attempts before the revision is rejected and the previous version is kept. Every accepted version is immutable, and each trial records the Lore and Craft versions it used.

\paragraph{Examples.}
Tab.~\ref{tab:supp-memory-example} shows one Lore item and one Craft tool, taken from the stored records. Both concern grasp verification and do not depend on the object or the scene. The Lore item, learned on Object-P task~2, corresponds to the failure mode in Fig.~2: a successful lift judged as a failure because re-segmentation returns nothing. The Craft tool, learned on Object-P task~0, turns grasp monitoring during lift and transport into a reusable function.

\section{Prompts}
\label{sec:supp-prompts}

\paragraph{Agent prompt.}
The prompts extend the CaP-X~\cite{liu2026capx} prompt and its multi-turn protocol, in which the agent replies with either a new code block or a completion declaration. The first turn uses the prompt in Fig.~\ref{fig:supp-prompt-agent}, followed by the API documentation of Tab.~\ref{tab:supp-api}, the capture-window documentation and, when memory is available, the active Lore and Craft documentation. \texttt{\{goal\}} is the LIBERO-PRO instruction.

\paragraph{Multi-turn prompt.}
Every later turn uses the prompt in Fig.~\ref{fig:supp-prompt-multi}, followed by the console history and the Witness reports.

\begin{figure}[t]
\centering
\begin{lrbox}{\suppcodebox}\begin{minipage}{\dimexpr\textwidth-2\fboxsep-2\fboxrule\relax}
\scriptsize
\begin{verbatim}
You are controlling a Franka Emika robot with the APIs described below.
Goal: {goal}

This is a persistent, multi-turn development session. You are not required to solve the entire task in one
code block. A turn may be used only to inspect the current observation, measure geometry, test one assumption,
or print diagnostics for the next turn. Prefer a small evidence-producing step over an unsafe full manipulation
when important facts are unknown.

The Python execution namespace persists for this trial. Variables, imports, helper functions, and cached
measurements defined by an earlier code block remain available to later blocks unless you overwrite them. The
next prompt also contains every earlier code block and the complete stdout/stderr history, so use that evidence
instead of starting over.

[...]

A special function is available:

    yield_to_next_turn(reason="", **diagnostics)

It immediately and successfully stops the current code block. It does NOT mean the task is finished. Use it
after an observation/diagnostic-only turn, when a precondition fails, or before an uncertain motion. Put concise
JSON-serializable measurements in diagnostics so the next turn can repair the correct stage.

[...]

Write robust code with explicit error handling and clear stopping conditions. Anticipate recoverable failures
and handle them locally when there is a justified recovery; otherwise end the current block early with
yield_to_next_turn(...) and preserve the evidence for the next turn. [...]

Print concise, clearly labelled information that will make the next repair actionable. [...] Never access
privileged simulator state or task-completion truth.
\end{verbatim}
\end{minipage}\end{lrbox}
\fcolorbox{black!35}{black!3}{\usebox{\suppcodebox}}
\caption{First-turn agent prompt (abridged; omissions marked [...]).}
\label{fig:supp-prompt-agent}
\end{figure}

\begin{figure}[t]
\centering
\begin{lrbox}{\suppcodebox}\begin{minipage}{\dimexpr\textwidth-2\fboxsep-2\fboxrule\relax}
\scriptsize
\begin{verbatim}
This is another step in the same persistent execution namespace.

All code blocks executed so far are shown below in execution order:
{executed_code}

You have multiple future turns available. You may:
- run a diagnostic-only block and call yield_to_next_turn(...),
- make a small targeted repair while preserving working definitions and stages,
- continue the manipulation from the current simulator state, or
- return FINISH only when the available non-privileged evidence indicates completion.

Read the complete console history and the latest Witness description included below. Diagnose the earliest failing
stage before changing code. Preserve successful stages. Do not restart or rewrite the whole solution without
evidence. Any new code should validate preconditions, stop safely on uncertainty, and print the measurements
needed by a later turn. [...]

Please respond with exactly one of:
- REGENERATE followed immediately by one complete Python code block in a fenced block;
- FINISH if the task appears complete.
\end{verbatim}
\end{minipage}\end{lrbox}
\fcolorbox{black!35}{black!3}{\usebox{\suppcodebox}}
\caption{Multi-turn prompt used at every later turn (abridged; omissions marked [...]).}
\label{fig:supp-prompt-multi}
\end{figure}

\paragraph{Updater prompts.}
The Lore and Craft updaters receive the trial record of Sec.~\ref{sec:supp-lorecraft}, the active library and the output schema. Their system prompts state the update rules of Sec.~\ref{sec:supp-lorecraft}. The Craft updater additionally receives the native API documentation and the capability inventory, so that it does not recreate an existing primitive. The full prompts are released with the code.

\section{Evaluation Protocol and Cost Accounting}
\label{sec:supp-protocol}

\paragraph{Benchmark.}
LIBERO-PRO~\cite{zhou2026liberopro} perturbs the Object, Goal and Spatial suites of LIBERO~\cite{liu2023libero}. Position perturbations (P) change initial object layouts, and task perturbations (T) change the instruction and target configuration. Each of the six splits has 10 tasks. Zero-shot evaluation uses the 50 provided initial states per task (seeds 1--50). For online adaptation (seeds 51--65), validation (seeds 66--75) and held-out evaluation of adapted programs (seeds 1--50), the seed samples the initial layout, following ASPIRE~\cite{lu2026aspire}.

\begin{table}[t]
\centering
\small
\caption{Model calls and cost of the 3{,}000 zero-shot rollouts, by call type. Tokens are in millions.}
\label{tab:supp-cost}
\begin{tabular}{@{}lrrrrrr@{}}
\toprule
Call type & Calls / rollout & Cache-hit in & Cache-miss in & Output & Cost (\$) & Share \\
\midrule
Witness: initial scene & 1.00 & 18.0 & 0.9 & 3.4 & 2.20 & 0.5\% \\
Agent: first code block & 1.00 & 16.5 & 1.5 & 4.4 & 2.92 & 0.7\% \\
Agent: later decisions & 16.24 & 958.5 & 528.6 & 467.9 & 362.92 & 90.5\% \\
Agent: malformed-response repair & 0.34 & 29.2 & 0.2 & 9.3 & 5.69 & 1.4\% \\
Witness: block start/end comparison & 14.48 & 263.8 & 36.6 & 27.3 & 22.67 & 5.7\% \\
Witness: capture-window frames & 1.76 & 29.7 & 7.6 & 5.5 & 4.51 & 1.1\% \\
\midrule
Total & 34.81 & 1{,}315.7 & 575.4 & 517.8 & 400.91 & 100\% \\
Per rollout & & & & & 0.134 & \\
\bottomrule
\end{tabular}
\end{table}

\paragraph{Success metrics.}
The environment success predicate is evaluated at every simulation step. A rollout counts as an \emph{ever-success} if the predicate holds at any step, and as a \emph{final success} if it holds when execution ends. Execution ends when the agent declares completion or the action-round or simulation-step budget is exhausted.

\paragraph{Program selection.}
After online trials 5, 10 and 15, and after the last successful online trial, the current program and memory are frozen as candidates. Each candidate is evaluated on validation seeds 66--70 without model calls. The candidate with the highest validation score is selected. If several candidates tie for the highest score, only those are evaluated on seeds 71--75, and the one with the highest total over seeds 66--75 is selected.

\begin{table}[t]
\centering
\small
\caption{Zero-shot execution statistics per split (50 seeds $\times$ 10 tasks per split). Disrupted: rollouts in which the goal state is reached and later undone; restored: those among them that end in final success.}
\label{tab:supp-zs-stats}
\begin{tabular}{@{}lccccccc@{}}
\toprule
 & Object-P & Object-T & Goal-P & Goal-T & Spatial-P & Spatial-T & All \\
\midrule
Final success (\%) & 59.2 & 53.2 & 54.0 & 17.2 & 51.6 & 62.2 & 49.6 \\
Ever-success (\%) & 60.0 & 55.2 & 62.0 & 26.8 & 54.0 & 65.4 & 53.9 \\
Goal disrupted (rollouts) & 24 & 40 & 143 & 73 & 77 & 123 & 480 \\
\quad restored (rollouts) & 20 & 30 & 103 & 25 & 65 & 107 & 350 \\
Agent turns / rollout & 14.3 & 16.2 & 18.6 & 23.6 & 15.6 & 15.1 & 17.2 \\
API calls / rollout & 240 & 473 & 552 & 687 & 378 & 308 & 440 \\
\bottomrule
\end{tabular}
\end{table}

\begin{table}[t]
\centering
\small
\caption{Zero-shot success (\%) with 95\% bootstrap intervals over seeds.}
\label{tab:supp-ci}
\begin{tabular}{@{}lccccccc@{}}
\toprule
 & Object-P & Object-T & Goal-P & Goal-T & Spatial-P & Spatial-T & Avg. \\
\midrule
Final & 59.2 {\scriptsize[55.6, 62.8]} & 53.2 {\scriptsize[49.0, 57.0]} & 54.0 {\scriptsize[50.4, 57.4]} & 17.2 {\scriptsize[14.0, 20.2]} & 51.6 {\scriptsize[48.0, 55.6]} & 62.2 {\scriptsize[58.4, 66.4]} & 49.6 {\scriptsize[48.0, 51.2]} \\
Ever & 60.0 {\scriptsize[56.4, 63.8]} & 55.2 {\scriptsize[51.2, 59.4]} & 62.0 {\scriptsize[58.4, 65.6]} & 26.8 {\scriptsize[23.2, 30.2]} & 54.0 {\scriptsize[50.0, 58.0]} & 65.4 {\scriptsize[61.6, 69.4]} & 53.9 {\scriptsize[52.2, 55.3]} \\
\bottomrule
\end{tabular}
\end{table}

\paragraph{Rollout count.}
The 34 target-task rollouts per task reported in the main paper are the 15 online trials plus, on average, about 19 validation rollouts. Held-out evaluation on seeds 1--50 is not counted, as for all methods in Tab.~2. Validation and held-out rollouts execute a frozen program and make no model calls.

\paragraph{Cost.}
Costs are computed from the token counts logged for every model call, at the off-peak list prices of DeepSeek for \texttt{deepseek-flash} (DeepSeek-V4.1-Flash-1017): \$0.003, \$0.15 and \$0.60 per million cache-hit input, cache-miss input and output tokens. Peak-hour prices are twice as high. Tab.~\ref{tab:supp-cost} breaks down the 3{,}000 zero-shot rollouts by call type. Later agent decisions account for 90.5\% of the cost, and all Witness calls together for 7.3\%.

\section{Baselines}
\label{sec:supp-baselines}

\paragraph{Published results.}
VLA results in Tab.~1 are taken from LIBERO-PRO~\cite{zhou2026liberopro} and the respective papers. The Opus~4.6 zero-shot row and all ASPIRE rows with Claude models are from ASPIRE~\cite{lu2026aspire} (arXiv v2; per-task results in Tables~10 and 11, rollouts and cost in Tables~20 and 21). CaP-Agent0 on LIBERO-PRO is from CaP-X~\cite{liu2026capx} (Appendix~J). RATs~\cite{zhang2026rats} is evaluated on 10 initial states per task instead of 50. RHO~\cite{elmaaroufi2026rho} evolves policy code over 63 generations on unperturbed LIBERO; the evaluated policy makes no model calls. ASPIRE split averages are computed from its per-task rows.

\paragraph{CaP-Agent0 on CaP-Bench.}
We run the released CaP-Agent0 implementation with DeepSeek-V4.1-Flash-1017 in place of its original model ensemble, keeping its nine-query ensemble with synthesis, its skill-library task configurations and its default budgets.

\paragraph{ASPIRE reproduction.}
We run ASPIRE's released harness and its iterative-debugging agent skill with the Claude Code CLI, using DeepSeek-V4.1-Flash-1017 as the model backend, under the CaAP budget. Evolutionary search and the cross-task skill library are not used, which corresponds to the debugging-only setting of ASPIRE in Tab.~2. Development uses seeds 51--65, each at most once. Program versions are then selected on validation seeds 66--75 with the remaining budget, for at most 34 episodes per task in total. The selected version is evaluated on seeds 1--50 without further model calls. The harness enforces the episode cap and the seed ranges.

\begin{figure}[t]
\centering
\includegraphics[width=\textwidth]{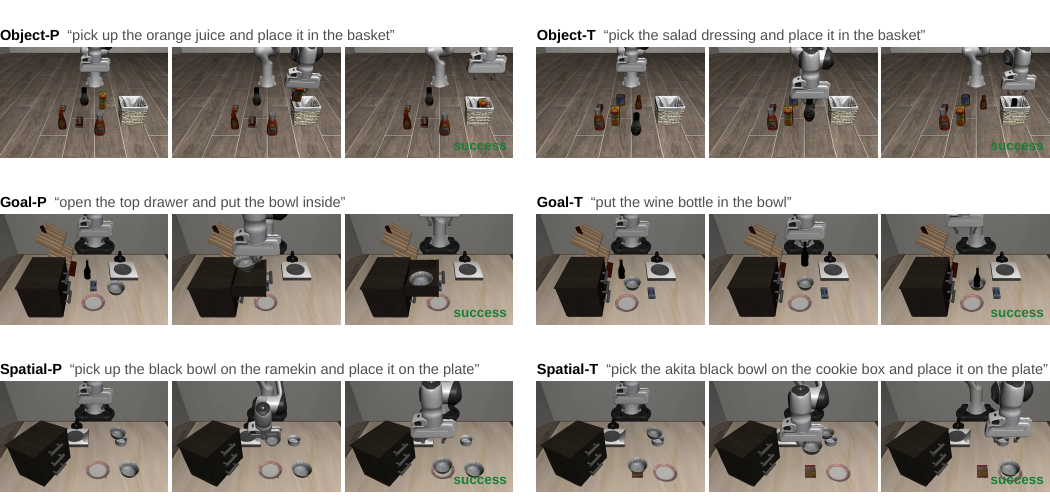}
\caption{Successful CaAP rollouts, one per LIBERO-PRO split, with the task instruction above each row of agent-camera frames (start, during manipulation, end).}
\label{fig:supp-qual}
\end{figure}

\begin{figure}[t]
\centering
\includegraphics[width=\textwidth]{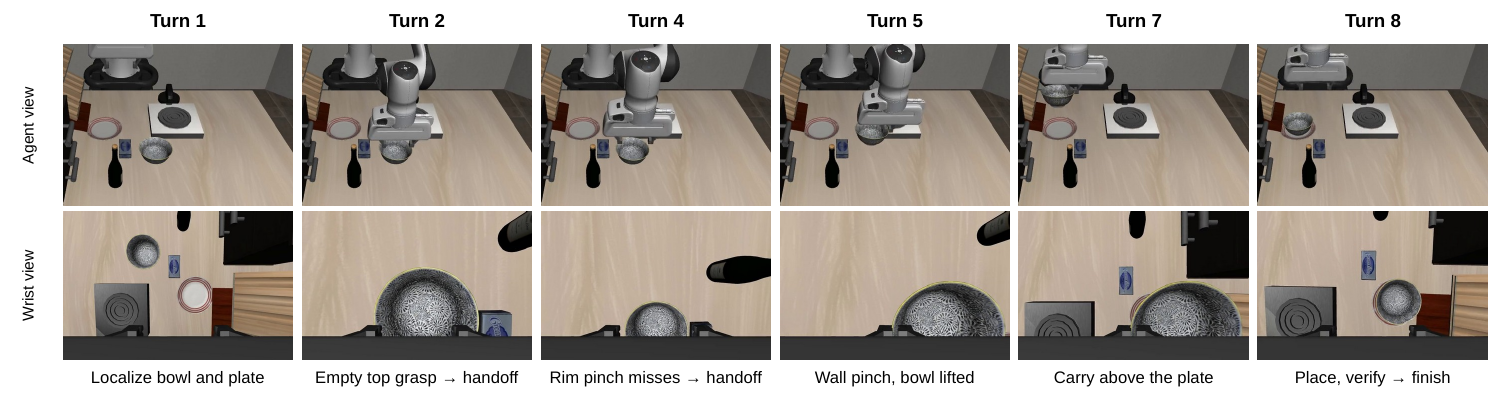}
\caption{Simulation frames of the case trial (Goal-P task~8, seed~41) during key turns, from the agent camera (top, cropped) and the wrist camera (bottom).}
\label{fig:supp-case}
\end{figure}

\section{Additional Zero-Shot Results}
\label{sec:supp-zeroshot}

\paragraph{Execution statistics.}
Tab.~\ref{tab:supp-zs-stats} gives per-split statistics of the 3{,}000 zero-shot rollouts. Goal-T rollouts use the most agent turns, consistent with it being the hardest split in Tab.~1. Of the 480 rollouts in which a later action disrupts an achieved goal state, 350 end in final success.

\paragraph{Confidence intervals.}
Tab.~\ref{tab:supp-ci} gives 95\% bootstrap intervals that resample the 50 seeds of each task with replacement.

\begin{table}[t]
\centering
\small
\caption{Turn-by-turn summary of the case trial. ``Code'' is what the generated code measured; ``Witness'' is the capture-window report the next turn received. Gripper width is normalized (1 = open).}
\label{tab:supp-case}
\begin{tabularx}{\textwidth}{@{}c>{\raggedright\arraybackslash}p{0.21\linewidth}>{\raggedright\arraybackslash}X>{\raggedright\arraybackslash}p{0.19\linewidth}@{}}
\toprule
Turn & Action & Code and Witness & Exit \\
\midrule
1 & Segment the bowl and the plate; measure their 3D geometry & Code: bowl and plate localized; centers and extents stored in \texttt{trial\_state}. & return \\
2 & Top-down grasp at the bowl center, then lift (keyframe window) & Code: the gripper closes to width 0.015, so nothing is between the fingers; it opens and retreats before lifting. Witness: the fingers straddle the rim and close, but the bowl stays on the table. & handoff: grasp appears empty \\
3 & Re-measure the bowl and plan a rim grasp; no motion & Code: bowl shifted by 6\,mm; rim grasp pose and pre-grasp computed. & handoff: rim-grasp plan ready \\
4 & Pinch the rim from the $+X$ side (keyframe window) & Code: width 0.016 after closing, the pinch missed. Witness: contact near the rim, but a secure pinch is uncertain. & handoff: $+X$ rim pinch missed \\
5 & Pinch the bowl wall along $Y$ and lift (keyframe window) & Code: width 0.070; grasp judged by re-segmentation (bowl center at 0.10\,m). Witness: the bowl does not visibly follow the gripper. & handoff: bowl grasped and lifted \\
6 & Resolve the disagreement before transport; no motion & Code: bowl points span 0.078--0.131\,m in the agent view, well above the table, so the bowl is lifted and not on the table. & handoff: lift verification computed \\
7 & Measure the held offset and carry the bowl above the plate (keyframe window) & Code: grasp offset 5.7\,cm; gripper width 0.065 after transport, still holding. & handoff: ready to lower and release \\
8 & Lower, release and retreat; verify placement (keyframe window) & Code: after release the bowl lies 3\,mm from the plate center at plate height, gripper open. Witness: the bowl rests within the plate rim, upright. & handoff: bowl resting on plate \\
9 & --- & The agent declares completion. & finish (success) \\
\bottomrule
\end{tabularx}
\end{table}

\section{Qualitative Results}
\label{sec:supp-qual}

Fig.~\ref{fig:supp-qual} shows one successful CaAP rollout on each LIBERO-PRO split, taken from online-adaptation trials (seeds 51--65). Frames are from the agent camera at the start of the trial, during manipulation and at the end. The examples cover pick-and-place into a basket under position and task perturbations (Object), a two-stage task that opens a drawer before placing the bowl inside and a changed target configuration (Goal), and bowls identified by spatial relations to other objects (Spatial). Sec.~\ref{sec:supp-case} follows a single trial turn by turn.

\section{Case Study: A Complete Trial}
\label{sec:supp-case}

We follow one zero-shot trial turn by turn: Goal-P task~8 (\emph{put the bowl on the plate}), seed~41, with empty memory. The trial succeeds after 9 agent turns, 8 code blocks and 5 of the 20 action rounds. Fig.~\ref{fig:supp-case} shows the scene at key turns and Tab.~\ref{tab:supp-case} summarizes every turn. Code and Witness reports below are quoted from the trial log; code is abridged (marked [...]).

\paragraph{Monitoring and handoff stop each failed grasp early.}
Every grasp attempt is wrapped in a capture window and followed by a check on the gripper width. In turn 2 the check fails right after closing, and the block hands off before lifting (Fig.~\ref{fig:supp-case-code}, top). The bowl is not disturbed, the namespace keeps the measured geometry, and each later turn changes only the grasp: a rim grasp in turns 3--4, then a pinch of the bowl wall along the other axis in turn 5, which succeeds.

\begin{figure}[t]
\centering
\begin{lrbox}{\suppcodebox}\begin{minipage}{\dimexpr\textwidth-2\fboxsep-2\fboxrule\relax}
\scriptsize
\textbf{Turn 2:} grasp check and handoff.
\begin{verbatim}
start_visual_recording(mode="keyframe", reason="top grasp and lift of the bowl")
open_gripper(); goto_pose(pre_grasp, q_down); goto_pose(grasp_pos, q_down); close_gripper()
closed_width = float(get_observation()["robot_cartesian_pos"][7])
# With the bowl between the fingers, the gripper should not close fully.
if closed_width < 0.15:
    open_gripper(); goto_pose(pre_grasp, q_down)
    end_visual_recording(reason="grasp closed fully; no bowl retained")
    trial_state["stage"] = "grasp_missed"
    yield_to_next_turn("grasp appears empty", closed_width=closed_width,
                       grasp_pos=grasp_pos.tolist(), trial_state=trial_state)
[...]
\end{verbatim}
\textbf{Turn 6:} opening comment of the verification block.
\begin{verbatim}
# Earliest failing stage to diagnose: Execution 5 reported "grasped & lifted" via
# segmentation (new_center z=0.0995), but Witness says the bowl did NOT follow the
# gripper upward. That is a segmentation false-positive risk (metallic gripper).
# Decide with a decisive test: min-Z of bowl points in the agent view.
#   - bowl on table  -> points near table top (z ~ 0)
#   - bowl lifted    -> all points raised well above table
# Corroborate with the wrist camera.
\end{verbatim}
\end{minipage}\end{lrbox}
\fcolorbox{black!35}{black!3}{\usebox{\suppcodebox}}
\caption{Code excerpts from the case trial, quoted from the trial log (abridged; omissions marked [...]; internal module names are replaced by those used in the paper).}
\label{fig:supp-case-code}
\end{figure}

\paragraph{Conflicting evidence is resolved before acting on it.}
After turn 5 the code reports a successful lift, while the Witness report states that the bowl does not visibly follow the gripper. Instead of transporting on either signal, turn 6 runs a dedicated test. Its opening comment is shown in Fig.~\ref{fig:supp-case-code} (bottom). The bowl points lie well above the table, so the agent proceeds. Because handoff returned control after each stage, this check costs one observation-only turn and no action round.

\paragraph{Completion follows a code check.}
In turn 8 the code lowers the bowl, releases it, and re-measures it: the bowl lies 3\,mm from the plate center at plate height with the gripper open, and the capture-window report agrees. The agent declares completion only after both checks. The verification steps in this trial are of the kind LoreCraft retains across trials, such as the Lore item and the \texttt{monitored\_lift} tool in Tab.~\ref{tab:supp-memory-example}.

\section{Real-Robot Details}
\label{sec:supp-real}

\paragraph{Platform.}
We use a bimanual mobile manipulator; the task uses one arm with its stock parallel gripper. A fixed third-person RGB-D camera provides the observations used for perception, and Witness receives frames from the third-person, wrist and head cameras. As in simulation, the agent, Witness and both LoreCraft updaters use DeepSeek-V4.1-Flash-1017, and commands and sensor streams are exchanged with the robot over rosbridge. Only the embodiment-specific interfaces are replaced: observation, arm and gripper control and camera access. The runtime, Self-Adaptive Process Monitoring, Witness, Heritable Handoff and LoreCraft are unchanged.

\enlargethispage{\baselineskip}
\paragraph{Task.}
The instruction is obtained from a human demonstration video: pour oil first, then add seasoning, and return each container upright to its original place after use. Each container sequence includes grasping, transport, pouring, returning upright and placement. The online run forms the program shown in Fig.~7.

\end{document}